\documentclass[conference]{IEEEtran}
\IEEEoverridecommandlockouts
\usepackage{multirow}
\usepackage{cite}
\usepackage{amsmath,amssymb,amsfonts}
\usepackage{algorithmic}
\usepackage{graphicx}
\usepackage{textcomp}
\usepackage{xcolor}
\usepackage{comment}

\usepackage{subcaption} 
\usepackage{authblk} 

\def\BibTeX{{\rm B\kern-.05em{\sc i\kern-.025em b}\kern-.08em
    T\kern-.1667em\lower.7ex\hbox{E}\kern-.125emX}}
\begin{document}

\title{Human-Centered Explainable AI for Security Enhancement: A Deep Intrusion Detection Framework\\
}

\author[1]{Md Muntasir Jahid Ayan\textsuperscript{*}\thanks{\textsuperscript{*} These authors have contributed equally to this work.}}
\author[2]{Md. Shahriar Rashid\textsuperscript{*}}
\author[3]{Tazzina Afroze Hassan}
\author[2]{Hossain Md. Mubashshir Jamil}
\author[1]{\\Mahbubul Islam}
\author[5]{Lisan Al Amin}
\author[4]{Rupak Kumar Das}
\author[6]{Farzana Akter}
\author[7]{Faisal Quader\textsuperscript{\textdagger}\thanks{\textsuperscript{\textdagger} Corresponding author.}}

\affil[1]{\textit{Department of Computer Science and Engineering, United International University (UIU),  Dhaka 1212, Bangladesh}}
\affil[2]{\textit{Department of Electrical and Electronic Engineering, Islamic University of Technology, Gazipur 1704, Bangladesh}}
\affil[3]{\textit{Department of Computer Science and Engineering (CSE), University of Asia Pacific (UAP), Dhaka 1207, Bangladesh}}
\affil[4]{\textit{College Of Information Sciences and Technology, Pennsylvania State University, University Park, PA 16802, USA}}
\affil[5]{\textit{Department of Information Systems, University of Maryland, Baltimore, 21250, Maryland, USA}}
\affil[6]{\textit{Department of Information Technology, Washington University of Science and Technology, Alexandria, VA }}
\affil[7]{\textit{College of Engineering and Information Technology, University of Maryland, College Park, 20742, Maryland, USA}}

\maketitle

\begin{abstract}
The increasing complexity and frequency of cyber-threats demand intrusion detection systems (IDS) that are not only accurate but also interpretable. This paper presented a novel IDS framework that integrated Explainable Artificial Intelligence (XAI) to enhance transparency in deep learning models. The framework was evaluated experimentally using benchmark dataset NSL-KDD demonstrating superior performance compared to traditional IDS and black-box deep learning models. The proposed approach combined Convolutional Neural Network (CNNs) and Long Short-Term Memory (LSTM) network for capturing temporal dependencies in traffic sequences. Our Deep Learning results showed that both CNN and LSTM reached 0.99 for accuracy whereas LSTM outperformed CNN at macro average precision, recall and F-1 score. For weighted average precision, recall and F-1 score, both model scored almost similar. To ensure interpretability, XAI model SHapley Additive exPlanations (SHAP) was incorporated, enabling security analysts to understand and validate model decisions. Some notable influential features were srv\_serror\_rate, dst\_host\_srv\_serror\_rate, and serror\_rate for both models as pointed out by SHAP. We also conducted a trust-focused expert survey based on IPIP6 and Big Five personality traits via an interactive UI to evaluate the system’s reliability and usability. This work highlighted the potential of combining performance and transparency in cybersecurity solutions and recommends future enhancements through adaptive learning for real-time threat detection.
\end{abstract}


\section{Introduction}
The roles of information and communication technology (ICT) systems are essential in every area of industry and human life~\cite{qazi2023hdlnids}. In recent years, many organization have been vulnerable due to cyber-attacks which led to the creation of Intrusion Detection System (IDS). It is a network security approach to detect cyber-attacks or malicious intrusion~\cite{hnamte2023novel}. Each cyber-attack results in financial loss, reputational damage also potential legal liabilities. The formation of IDS has an effect on both academic section and business sector in a global scale~\cite{hnamte2023novel}.
Intrusion detection has always been a significant area of study in network security since it is critical to recognize anomalous access to protected internal networks~\cite{awajan2023novel}. To preserve trust and privacy in these systems, human-centered explainability of underlying AI decisions becomes essential.
As cyber-attacks grow in complexity, there is demand of solutions which are sophisticated and adaptive to protect crucial infrastructure and sensitive information. Network Intrusion Detection Systems (NIDS) are in the frontline of these efforts, functioning as a vital role of defense against malicious activities. However, the rapid evolution of cyber threats necessitates ongoing advancements in detection techniques to ensure both accuracy and reliability. 
In recent years, machine learning has emerged as a tool to address cyber security which has the ability to detect different patterns of attacks and work through it~\cite{hnamte2023dcnnbilstm, kasongo2023deep}. Despite this, several challenges remain unaddressed, particularly in the context of network intrusion detection. Existing machine learning methods, such as decision trees, random forests, and basic neural networks, have showed moderate success in identifying intrusions. In some researches Convolutional Neural Network (CNN) has been deployed to detect attacks. These advanced models are complex and hard to understand and as a result, often their decision making processes are questioned. Explainable AI (XAI) aims to explain these decision making processes. This research aims to enhance the performance of NIDS by adopting a multilayered feature extraction mechanism and incorporating Explainable Artificial Intelligence (XAI) to bridge the gap between detection efficacy and interpretability. This study is aiming to integrate cutting-edge machine learning techniques, particularly Convolutional Neural Networks (CNNs) and Long Short-Term Memory (LSTM) networks, to process different types of features from network data~\cite{du2023nids}. The research covers on topics like deep learning, temporal data analysis, and cybersecurity. Additionally, for transparency in its decision-making process there is incorporation of XAI which is highly effective for intrusion detection. XAI methods are a set of techniques that help users understand how AI models make decisions. XAI methods can help identify potential biases, improve model accuracy, and build trust in AI models. 
To tackle the current challenges, this study introduces an innovative approach to network intrusion detection by utilizing CNNs for extracting categorical features and LSTMs for analyzing temporal features. Additionally, the research incorporates XAI to improve the model’s interpretability, enabling security professionals to comprehend the rationale behind detection outcomes. This approach seeks to balance accuracy and clarity, providing a complete solution to today’s network security issues. The principle contributions of this article are the following:
\begin{itemize}
\item We discuss and apply a comprehensive architectural overview of an NIDS system based on DL models, namely CNN and LSTM.
\item We apply XAI, specifically SHapley Additive exPlanations (SHAP) to get an explanation of the DL system to better understand its decisoin making process. 
\item We implement an UI system to verify the trustworthiness, reliability and usability of the NIDS system based on the SHAP explanations.
\item We also discuss some challenges and research directions regarding the application of XAI in NIDS.

\end{itemize}
The rest of the article is illustrated as follows: Section \ref{section_LR} provides a brief summary and contribution of existing literature.
Section \ref{section_method} presents an overview of the whole project along with short description of the DL and XAI models, dataset, performance evaluation metrics used in the project. In section \label{section_resultanalysis}, the DL and XAI model's outputs are analyzed. Finally, we conclude our work and discuss some future research directions in section \ref{section_conc_fut}. 

\begin{figure*}[htbp]
\centerline{\includegraphics[scale=.7]{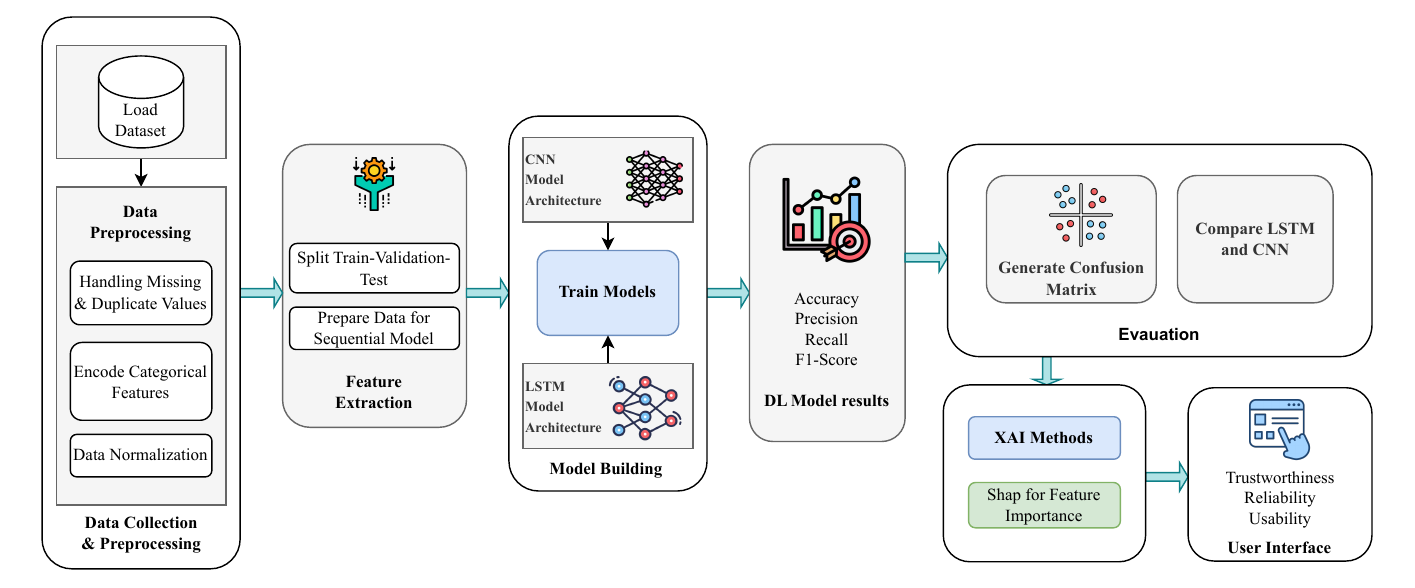}}
\caption{Our Proposed Framework for Trustworthy and Explainable Deep Learning-Based Network Intrusion Detection System and User Interface Driven Expert Verification Method.}
\label{fig:model}
\end{figure*}

\section{Related Work}\label{section_LR}

Network Intrusion Detection Systems (NIDS) are critical security mechanisms that monitor network traffic to identify malicious activities and policy violations. Relying on manually created signatures, early NIDS solutions were mostly rule-based or signature-driven, which offered lightweight and deployable intrusion detection~\cite{roesch1999snort}. While effective against known attacks, such systems struggled with zero-day threats and generated high false positives due to their static nature. To address these shortcomings, researchers gradually moved towards anomaly detection and machine learning approaches, enabling the system to detect unseen attacks by learning statistical and behavioral patterns in network traffic~\cite{wei2019design}. The development of machine learning (ML) methods led to improvements in generalization and accuracy. Support vector machines, decision trees, and random forests have all been extensively studied because they provide better detection rates and flexibility~\cite{adewole2025intrusion}. To further optimize performance, feature engineering and optimization strategies were introduced. For instance, one work applied a genetic algorithm–based feature selection method with a hypergraph and SVM classifier, achieving high detection accuracy while reducing feature redundancy in the NSL-KDD dataset~\cite{raman2017efficient}. Similarly, dimensionality reduction approaches such as Principal Component Analysis (PCA) were used to lower computational costs on large datasets. Despite these improvements,, these models often lacked scalability for high-speed networks and exhibited weaker interpretability compared to rule-based methods. More recently, deep learning (DL) methods have received more popularity in NIDS research due to their ability to automatically extract hierarchical features from raw traffic. While Recurrent Neural Networks (RNNs) and Long Short-Term Memory (LSTM) architectures efficiently model sequential dependencies in connection records, Convolutional Neural Networks (CNNs) have been utilized for capturing spatial traffic patterns. Hybrid approaches like CNN-LSTM and CNN-GRU achieved stronger detection performance with prominent datasets like NSL-KDD and CIC-IDS2017 ~\cite{pandey2025lightweight, mehmmod2025erbm}. Despite high accuracy and reduced false positives, these models are computationally intensive and, more importantly, suffer from the “black-box” problem, where stakeholders find it difficult to understand how decisions are made~\cite{adewole2025intrusion}. This dilemma of trust and transparency has led to the incorporation of explainable artificial intelligence (XAI) into IDS research. Explainability tools like SHAP and LIME provide post-hoc explanations by revealing the most influential features contributing to a given prediction, thereby making deep learning models more interpretable to security analysts. While several studies have employed rule-induction or interpretable ML frameworks~\cite{adewole2025intrusion, mehmmod2025erbm}, few works have applied modern XAI methods directly to deep neural architectures in the NIDS domain. Furthermore, there is a gap of understanding whether security professionals can trust, interpret, and rely on such outputs in practical contexts because the human-centered effectiveness of these explanations is rarely validated by the explainability studies. Despite these advancements, the gap still remains in combining the reliability needed for real-world deployment with the high performance of CNN and LSTM-based NIDS. Our study addresses this by using SHAP and LIME to offer comprehensible explanations for predictions generated by a CNN and Bi-LSTM NIDS trained on NSL-KDD. Beyond technical evaluation, we further developed a dedicated user interface (UI) to collect cognitive feedback from industry experts on the trust, reliability, and usability of the generated explanations. This unique inclusion of expert validation bridges the gap between algorithmic explainability and human trust in XAI-driven NIDS, which advances the field towards deployable, user-centered security solutions.

\section{Methodology}\label{section_method}
\subsection{Project Overview}\label{subection_overview}
 In our research we have used NSL-KDD dataset as benchmark dataset. NSL-KDD dataset has 43 columns along with 125973 rows. There are 23 types of attack on target class, from those target class we mapped into 5 classes: 'Normal', 'Dos', 'R2L', 'Probe', and 'U2R' with the help of security domain expert and survey result. Data distribution 67343, 45927, 11647, 995, 52 for Normal, Dos, R2L, Probe and U2R respectively. Dataset contains both categorical and numerical data. For categorical data we have used label encoding and used min-max scaling for scale the data. We have used 80\% data for training and 20\% for testing. In training dataset we have performed CNN and LSTM model for multiclass classification. Each model performed 50 epochs with batch size 64. 3 layered CNN model gives 99\% accuracy with softmax activation function, adam optimizer and sparse categorical crossentropy. On other side, 3 layered lstm with dropout rate 0.3, activation softmax, optimizer adam and categorical crossentropy also provides 99\% accuracy on multiclass classification. Both model provide 99\% accuracy. To better understand the model performance, we have explored explainable AI. Mostly ML and DL models are black box. XAI helps us to understand the models behavior. We performed SHAP on both models CNN and LSTM for better understanding of their multiclass classifications. After that we include conducted a study  to evaluate user perceptions of the proposed system, focusing on trust, reliability, and usability. Incorporating Big5 personality assessments and validated constructs provided insights into how individual differences influence system interaction. Figure \ref{fig:model} showed a graphical interpretation of our proposed framework of the overall system consisting of data preprocessing, feature extraction, DL model building, result evaluation, SHAP explanation and the UI testing.

\subsection{Model Description}\label{subsection_description}
\subsubsection{Convolutional Neural Network (CNN)}\label{ssubsection_cnn}
A CNN consists of three main layers: the input layer, the hidden layer, and the output layer. The input layer receives normalized array data. The hidden layer is comprised of convolutional, pooling, and fully connected layers. The convolutional layer utilizes convolutional kernels to extract features and map excitations. Once feature extraction is performed, the resulting feature maps are passed to the pooling layer. The pooling layer is responsible for feature selection and information filtering. It performs down sampling, sparsely processing feature maps to reduce data computation, significantly decrease parameter complexity, and mitigate over fitting. The fully connected layer, typically positioned at the end of the CNN, helps retain crucial feature information while minimizing loss. Finally, the output layer generates classification results for the extracted features~\cite{lecun2002gradient}.

\subsubsection{Long Short-Term Memory (LSTM)}\label{ssubsection_lstm}
In an LSTM network, data moves exclusively forward, ensuring that the state at time t depends solely on the information from previous time steps~\cite{hochreiter1997long}. Traditional RNNs suffer from the long-term dependency problem, where past information gradually fades as the number of neurons increases. LSTMs address this issue by preserving important information across all LSTM units using a mechanism called the memory cell, which functions like a conveyor belt~\cite{mendez2023long}. Each LSTM unit comprises a memory cell and three gates that control information flow by determining what to retain and what to discard. This mechanism enables LSTMs to learn long-term dependencies, effectively addressing the long-term dependency problem. The three gates, known as the forget gate, input gate, and output gate, play a crucial role in this process.

\subsubsection{SHapley Additive exPlanations (SHAP)}\label{ssubsection_shap}
SHAP is a technique that measures the importance of the input features of the DL model in order to explain its results by quantifying the impacts via Shapley values of the conditional expectation function of the original model. Later, it calculates the Additive Feature Attribution Methods to accumulate the influence of each feature to explain the overall results. Within the Additive Feature Attribution Methods, a unique solution satisfies three properties:  Local Accuracy, Missingness, and Consistency. Shapley values can meet these requirements. SHAP can average all possible non-linear or non-independent inputs to generate explanations~\cite{chen2023overview}.

\begin{figure*}[!h]
\label{fig:curves}
\centering
  \begin{subfigure}[b]{0.3\textwidth}
    \includegraphics[width=\textwidth]{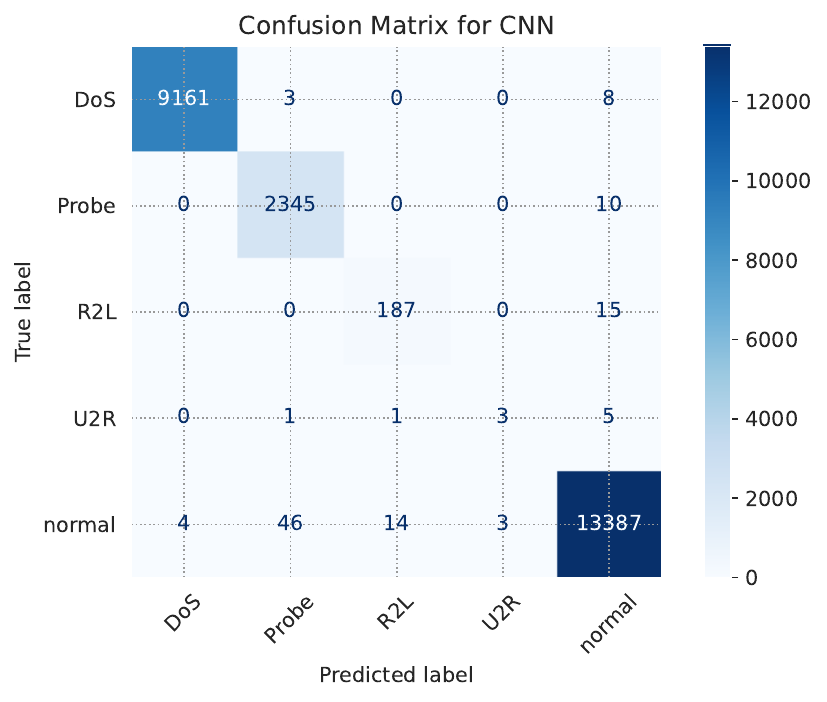}
    \caption{Confusion Matrix for CNN}
    \label{fig:cnncm}
  \end{subfigure}
  \begin{subfigure}[b]{0.3\textwidth}
    \includegraphics[width=\textwidth]{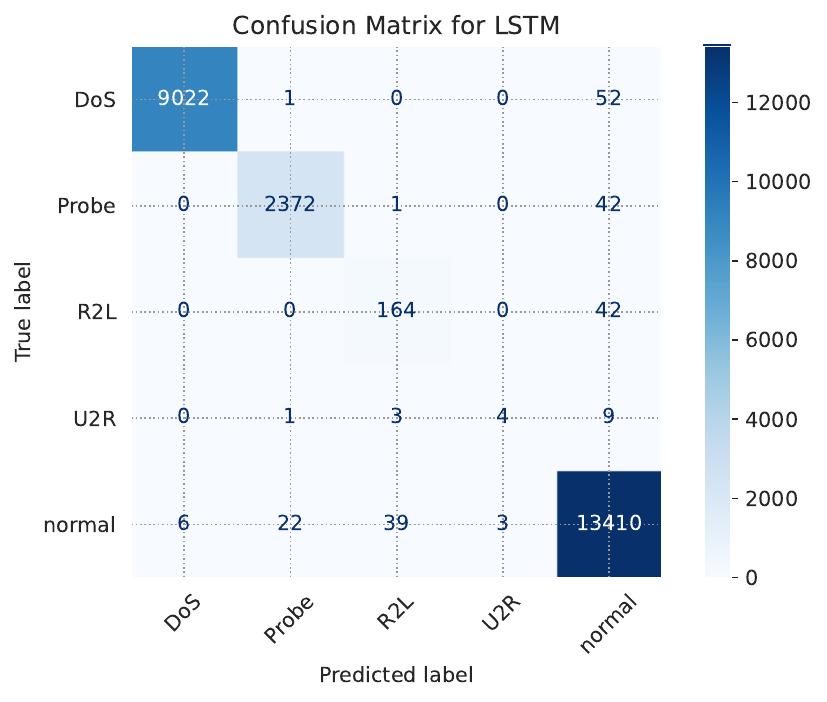}
    \caption{Confusion Matrix for LSTM}
    \label{fig:lstmcm}
  \end{subfigure}
  
  \begin{subfigure}[b]{0.3\textwidth}
    \includegraphics[width=\textwidth]{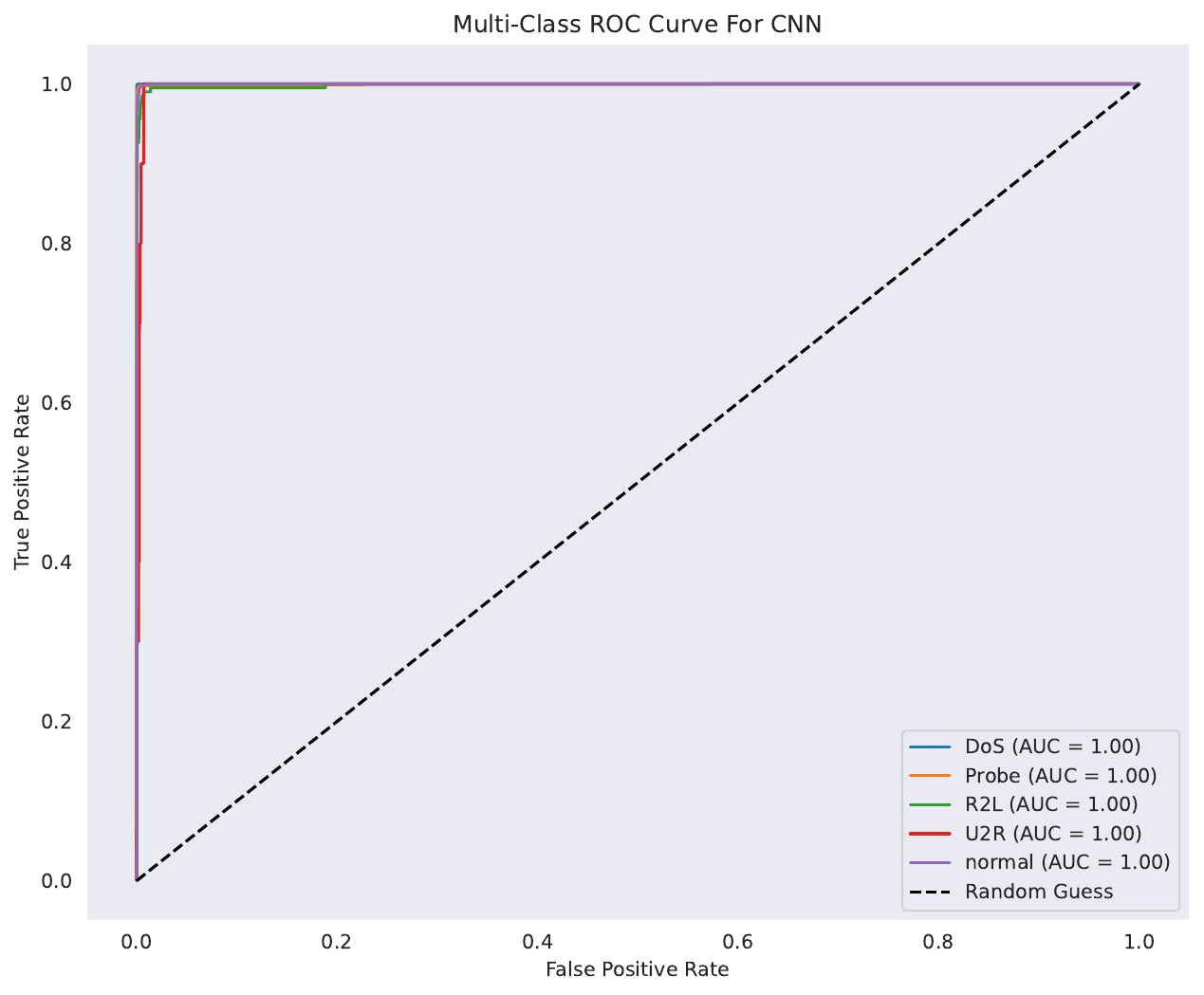}
    \caption{ROC Curve for CNN}
    \label{fig:cnnroc}
  \end{subfigure}
  \begin{subfigure}[b]{0.3\textwidth}
    \includegraphics[width=\textwidth]{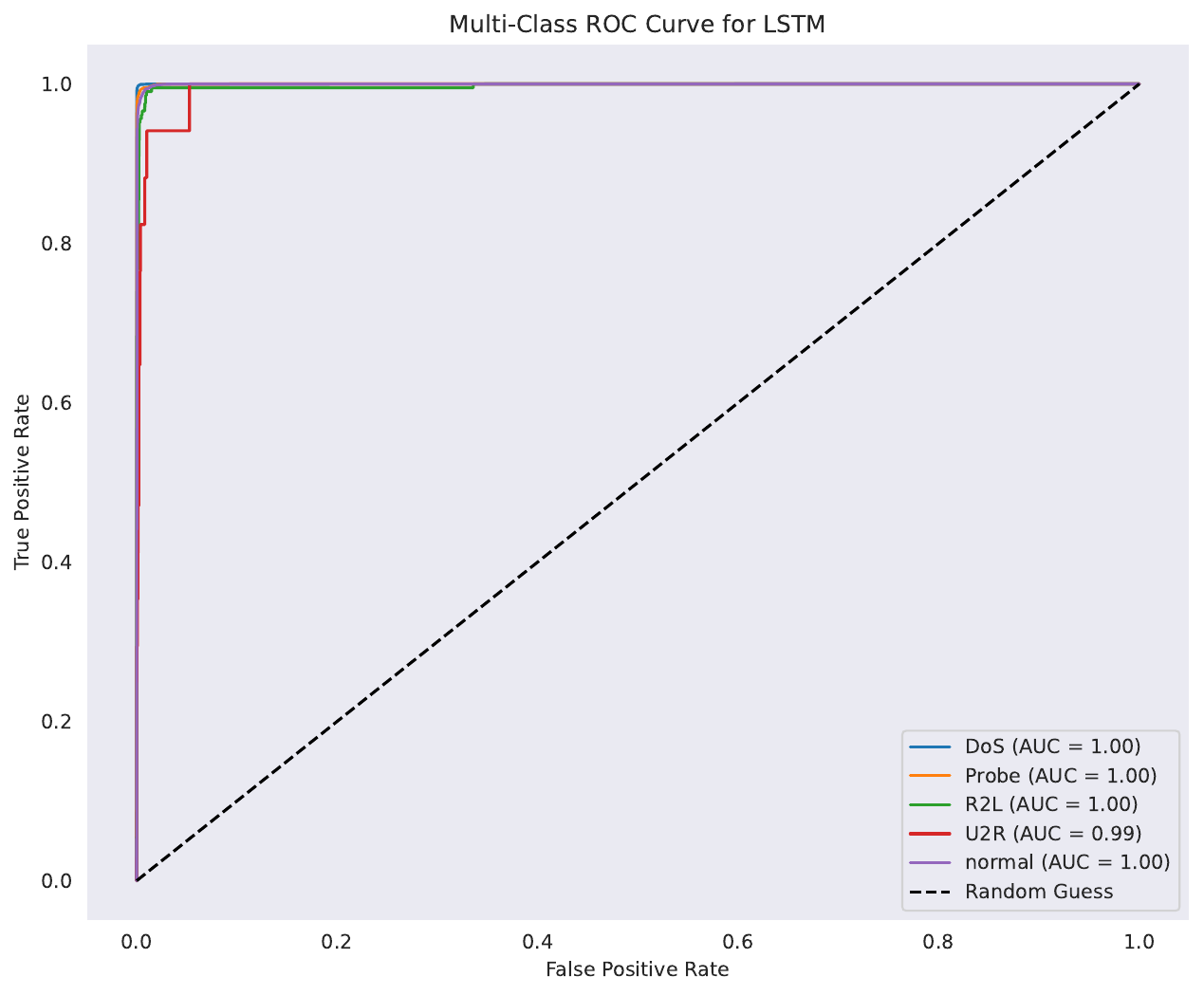}
    \caption{ROC Curve for LSTM}
    \label{fig:lstmroc}
  \end{subfigure}
  \caption{Deep Learning Model Results for Intrusion Detection: (a) Confusion Matrix for CNN, (b) Confusion Matrix for LSTM, (c) ROC Curve for CNN, and (d) ROC Curve for LSTM.}
\end{figure*}

\subsection{Dataset Description}\label{subsection_dataset}
This study utilized the NSL-KDD dataset~\cite{revathi2013detailed}, a refined version of the KDD Cup 99 benchmark widely used in intrusion detection research~\cite{tavallaee2009detailed}. The dataset comprises 41 features categorized into basic features (e.g., duration, protocol\_type), content-based features (e.g., logged\_in, num\_compromised), and traffic-based features (e.g., dst\_host\_srv\_count, dst\_host\_same\_src\_port\_rate). Each record is labeled as either normal or one of several types of attack, which we further grouped into five broad categories for multi-class classification: normal, DoS, Probe, R2L, and U2R. The complete dataset was split into training and testing sets in an 80:20 ratio~\cite{awad2023improved}, resulting in 100,771 training samples and 25,193 testing samples.

\subsection{Preprocessing and Feature Engineering}\label{subsection_preprocess}
Data preprocessing was an essential step to ensure quality and model compatibility. The dataset was first scanned for missing values, duplicates, and inconsistent entries, which were cleaned to enhance data integrity. Numerical features were normalized using min-max scaler, scaling all values to the range [0, 1] to prevent bias from features with larger numeric ranges. Categorical attributes such as protocol\_type, service, and flag were label-encoded into numerical values to maintain their categorical structure while allowing them to be used in the neural network models. The target variable was also label-encoded to represent the five target classes as integers from 0 to 4 respectively such as class 0 as DoS, class 1 as Probe, class 2 as R2L, class 3 as U2R, and class 4 as Normal. To accommodate the input requirements of the neural network models, feature matrices were reshaped: CNN inputs were reshaped to (samples, 41, 1) to treat the 41 features as one-dimensional sequences, while LSTM inputs were reshaped to (samples, 1, 41) to model the sequence along the temporal dimension. Rather than reducing dimensionality through feature selection methods such as Random Forest or PCA, we retained all 41 original features to allow deep learning models to learn patterns directly from raw data. This approach leveraged the strength of CNN in capturing spatial relationships between features, and LSTM’s ability to model dependencies over sequences. Features like src\_bytes, dst\_host\_same\_src\_port\_rate, wrong\_fragment, and logged\_in which frequently appeared in prior studies as critical indicators of intrusion—were preserved and allowed to be learned organically by the models. Although not explicitly transformed or engineered, the features were made model-ready through encoding, normalization, and reshaping.

\subsection{Performance Evaluation Metrics}\label{ssection_metrics}

The confusion matrix~\cite{townsend1971theoretical}is a fundamental evaluation tool in network intrusion detection systems (NIDS), enabling the assessment of classification~\cite{zhang2021new} performance across different traffic types. It quantifies the number of true positives (TP), true negatives (TN), false positives (FP), and false negatives (FN), where TP measures Intrusions correctly identified as intrusions, TN measures Normal traffic correctly identified as normal, FP measures Normal traffic incorrectly classified as intrusion (false alarm) lastly FN measures intrusions missed and classified as normal.
From this matrix~\cite{heydarian2022mlcm}, key evaluation metrics such as accuracy, which measures the overall correctness of the model by evaluating the proportion of total correctly predicted instances, precision indicates the proportion of correctly identified intrusion instances among all instances predicted as intrusions, recall measures the proportion of actual positives that were correctly identified, F1-score the harmonic mean of precision and detection rate, used to balance both false positives and false negatives. These indicators are critical for evaluating the trade-off between detection performance and false alarm rates in real-world NIDS deployment.


\begin{figure*}[!tb]
\centering
  \begin{subfigure}[b]{0.4\textwidth}
    \includegraphics[width=\textwidth]{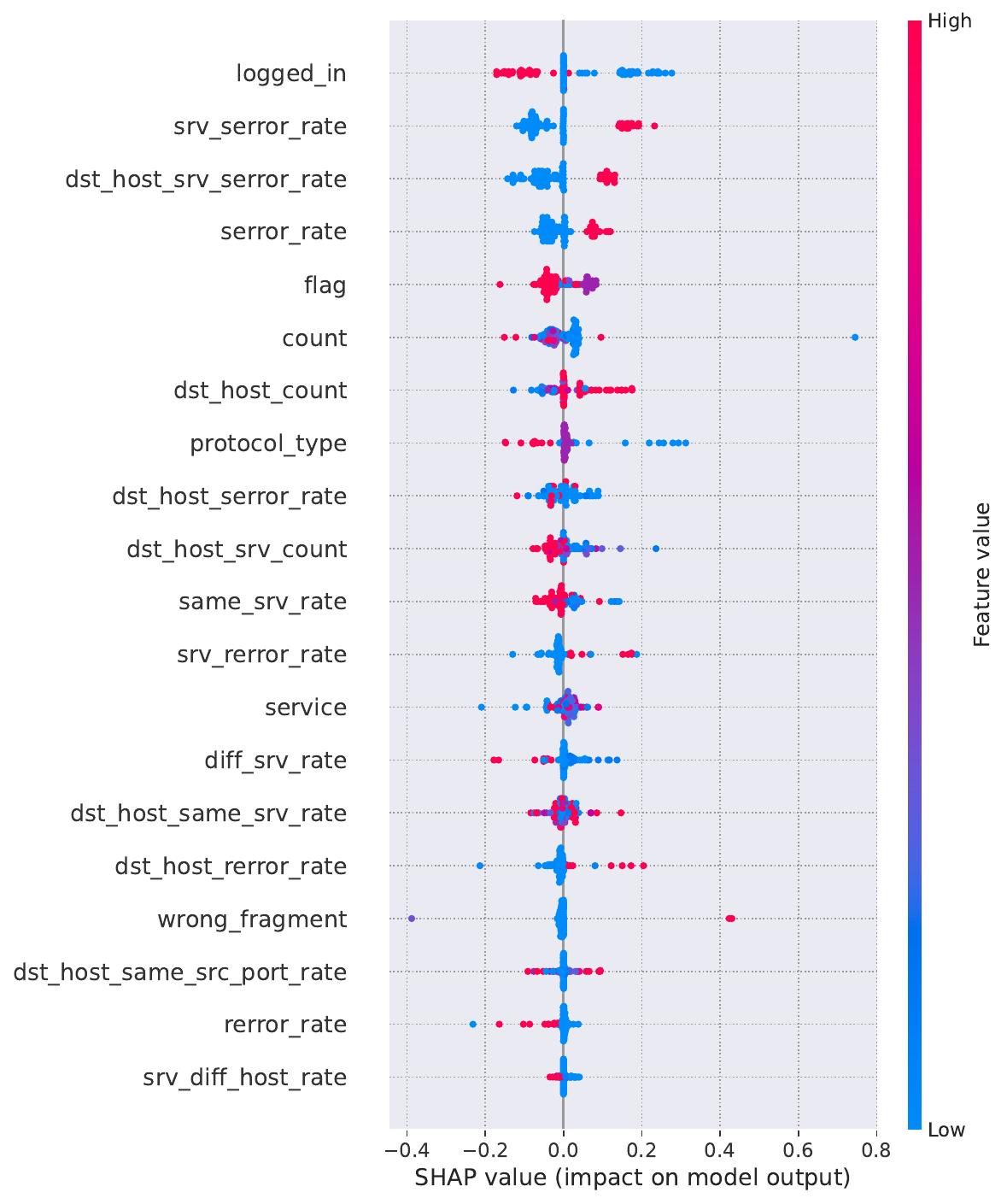}
    \caption{SHAP Results for CNN}
    \label{fig:cnnshap}
  \end{subfigure}
  \begin{subfigure}[b]{0.4\textwidth}
    \includegraphics[width=\textwidth]{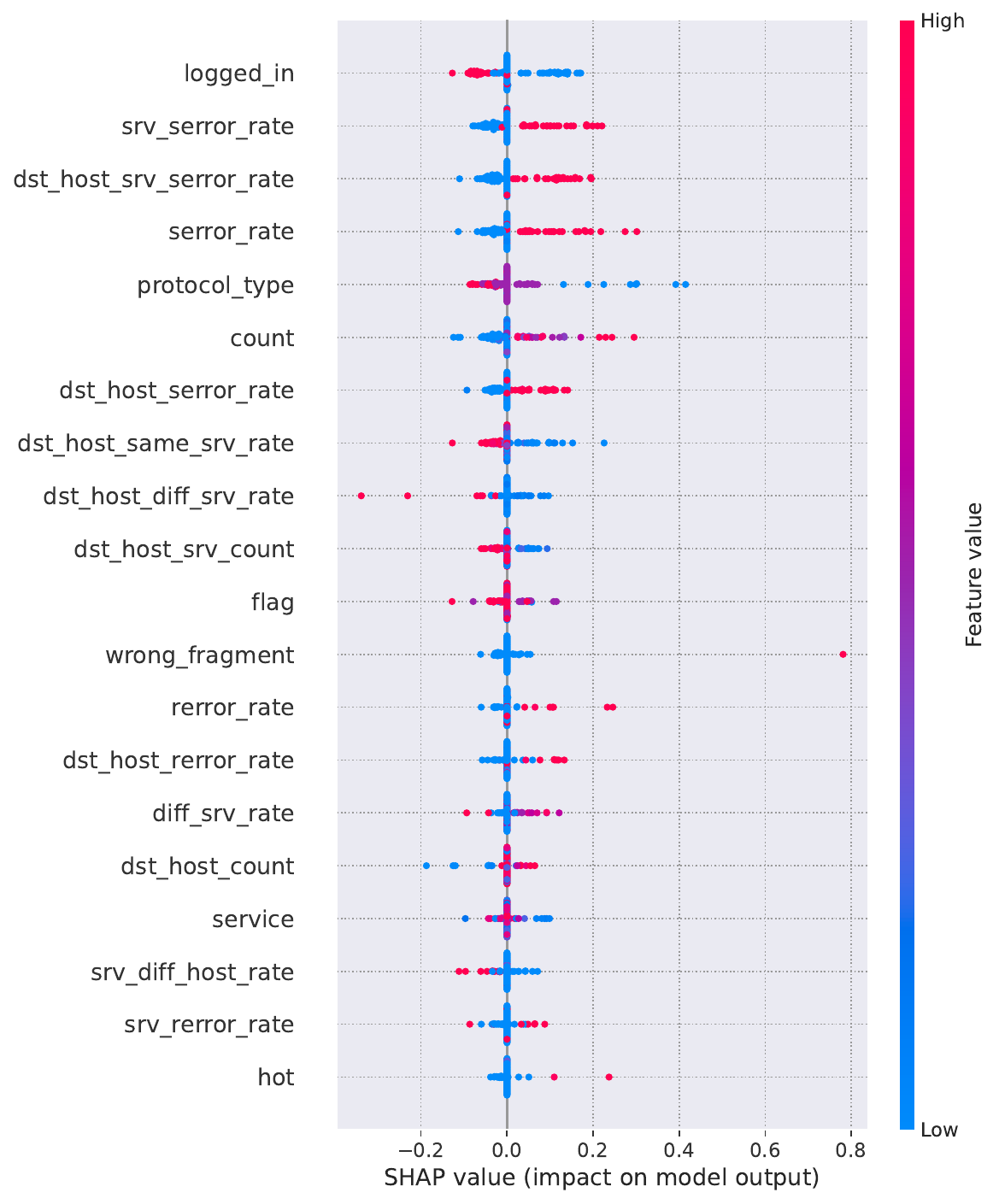}
    \caption{SHAP Results for LSTM}
    \label{fig:lstmshap}
  \end{subfigure}
  \caption{SHAP Results for Deep Learning Models: (a) SHAP Results for CNN, (b) SHAP Results for LSTM.}
\end{figure*}

\section{Result Analysis}\label{section_resultanalysis}

\begin{table*}[t]
\centering
\caption{Performance analysis of deep learning models based on accuracy, precision, recall and F1-score}
\label{tab:DLperformance}
\begin{tabular}{|c|c|c|ccc|ccc|}
\hline
\multirow{2}{*}{Dataset} & 
\multirow{2}{*}{Model} & 
\multirow{2}{*}{Accuracy} & 
\multicolumn{6}{c|}{Performance} \\ \cline{4-9}

& & &
\multicolumn{3}{c|}{Macro Avg} &
\multicolumn{3}{c|}{Weighted Avg} \\ \cline{4-9}

& & &
Precision & Recall & F1-score &
Precision & Recall & F1-score \\ \hline

\multirow{2}{*}{NSL-KDD} &
CNN &
0.99 &
0.88 & 0.84 & 0.86 &
0.99 & 0.99 & 0.98 \\ \cline{2-9}

& LSTM &
0.99 &
0.99 & 0.89 & 0.93 &
0.99 & 0.99 & 0.99 \\ \hline

\end{tabular}
\end{table*}


\subsection{CNN and LSTM Model Performance}\label{subsection_DLperformance}

Two deep learning models—a CNN and a LSTM network—were trained and evaluated on the NSL-KDD dataset for intrusion detection. CNN and LSTM outperforms in intrusion detection \cite{11022010} and both models achieved an overall accuracy of 99\%, demonstrating strong performance in distinguishing between normal and malicious network traffic showed in table \ref{tab:DLperformance}. The table \ref{tab:DLperformance} presents the overall performance analysis of the DL models. However, a closer examination of the evaluation metrics reveals some key differences in their behavior across different attack categories.

In terms of macro average scores, the CNN achieved a precision of 0.88, recall of 0.99, and F1-score of 0.86. The LSTM, on the other hand, achieved higher macro average precision at 0.99, but lower recall at 0.89 and an F1-score of 0.93. For weighted average metrics, both models performed similarly, with the CNN scoring 0.99 precision, 0.99 recall, and 0.98 F1-score, while the LSTM achieved 0.99 across all three metrics—precision, recall, and F1-score.

\begin{table}[]
\centering
\caption{Accuracy Per Attack Type}
\label{tab:accuracyperattacktype}
\begin{tabular}{llrrr}
\multicolumn{1}{c}{\textbf{Model}} &
  \multicolumn{1}{c}{\textbf{Attack Type}} &
  \multicolumn{1}{c}{\textbf{Precision}} &
  \multicolumn{1}{c}{\textbf{Recall}} &
  \multicolumn{1}{c}{\textbf{F1-Score}} \\
  \hline
  \\
  
CNN    & DOS      & 0.99   & 0.99   & 0.99   \\
       & Probe    & 0.97   & 0.99   & 0.98   \\
       & R2L      & 0.92   & 0.92   & 0.92   \\
       & U2R      & 0.5    & 0.33   & 0.4    \\
       & Normal   & 0.99   & 0.99   & 0.99   \\\\
         \hline
         \\
LSTM   & DOS      & 0.99   & 0.99   & 0.99   \\
       & Probe    & 0.98   & 0.98   & 0.98   \\
       & R2L      & 0.79   & 0.79   & 0.79   \\
       & U2R      & 0.57   & 0.23   & 0.33   \\
       & Normal   & 0.98   & 0.99   & 0.99  
\end{tabular}
\end{table}

When performance is broken down by attack type in table \ref{tab:accuracyperattacktype}, both models excelled in detecting Denial of Service (DOS) and Probe attacks, achieving near-perfect precision, recall, and F1-scores. The CNN scored 0.99 for all three metrics on DOS and 0.97, 0.99, and 0.98 for precision, recall, and F1-score respectively on Probe attacks. Similarly, the LSTM scored 0.99 across all metrics for DOS and 0.98 for each metric on Probe. For Remote to Local (R2L) attacks, the CNN outperformed the LSTM with a precision, recall, and F1-score of 0.92, while the LSTM achieved 0.79 across the board. In the case of User to Root (U2R) attacks, both models struggled, with the CNN scoring 0.50 precision, 0.33 recall, and 0.40 F1-score, and the LSTM performing slightly worse in recall and F1-score at 0.23 and 0.33, despite a higher precision of 0.57. For Normal traffic, both models performed very well, with the CNN scoring 0.99 in all metrics and the LSTM achieving 0.98 precision, 0.99 recall, and 0.99 F1-score.
Overall, while both models achieved high accuracy, the CNN demonstrated stronger recall and more balanced performance across difficult attack types like R2L and U2R, whereas the LSTM showed higher precision on average but struggled with minority class recall, particularly in U2R detection. Figure \ref{fig:cnncm}, \ref{fig:lstmcm}, \ref{fig:cnnroc}, and \ref{fig:lstmroc} showed the CNN and LSTM models' generated detection results in confusion matrix and ROC curve formats.

\subsection{Explainable AI Performance}\label{subsection_XAIperformance}

To interpret the decisions made by the models, we employed SHAP. SHAP was intended to provide global and local explanations; however, compatibility issues between TensorFlow 2.18.0 and SHAP 0.47.1 led to several challenges. Specifically, we encountered incorrect shap\_values dimensions and attribute errors in both DeepExplainer and GradientExplainer, despite attempts to reduce subset sizes and restructure model input/output layers. Despite this, Our goal is explaining the prediction of an instance by computing the contribution of each feature of SHAP. Figure \ref{fig:cnnshap} and \ref{fig:lstmshap} showed the explanation and influence of several features used by the CNN and LSTM models to provide insight of their working process.

\subsubsection{SHAP results for CNN}\label{ssubsection_CNNSHAPperformance}
To enhance the interpretability of the convolutional neural network (CNN) model, SHAP was utilized to analyze the contribution of input features to model predictions. SHAP is a unified framework for explainability rooted in cooperative game theory and provides both local and global interpretability. Given the three-dimensional input structure of the CNN model - specifically of shape samples, features - the input data were reshaped into two dimensions to comply with SHAP’s visualization requirements. A subset of 100 test instances was selected for explanation, and a representative background set of 100 samples from the training data was used to compute SHAP values efficiently.

Initially, SHAP's DeepExplainer was employed to leverage the internal architecture of the CNN model for generating feature attributions. In cases where DeepExplainer was incompatible with certain TensorFlow configurations, a fallback to GradientExplainer was applied to ensure the robustness of the interpretability process. The SHAP summary plot (figure - \ref{fig:cnnshap}) presented for class 0, corresponding to the Denial of Service (DoS) attack class, provides a global interpretation of how various input features influence the model's output toward identifying DoS traffic. The Y-axis of the plot lists the features ranked by their average absolute SHAP values, indicating overall importance, while the X-axis displays the SHAP value distribution for each feature, signifying its contribution to increasing or decreasing the model’s confidence in predicting the DoS class. The color gradient across the plot denotes the magnitude of feature values—ranging from low (blue) to high (red)—for each data instance.
The feature srv\_serror\_rate emerged as the most influential in predicting DoS attacks. High values of this feature (red points) were associated with strongly positive SHAP values, meaning that frequent server connection errors significantly increased the model's likelihood of classifying the traffic as DoS. Similarly, features such as dst\_host\_srv\_serror\_rate, serror\_rate, and dst\_host\_serror\_rate also showed substantial positive influence when their values were high, further indicating that high rates of connection failures and service errors are characteristic of DoS attack patterns. These features are consistent with typical DoS behaviors that involve overwhelming target hosts with connection requests, leading to high failure and error rates.
In contrast, the logged\_in feature showed an inverse relationship: low values (i.e., when users were not logged in) contributed positively toward DoS prediction, suggesting that attacks are generally launched without authenticated sessions. Other features such as protocol\_type, count, and flag had moderate influence, with their effect dependent on specific value combinations. Features like wrong-fragment and rerror\_rate, while included in the model, had relatively smaller SHAP values and thus contributed less to the CNN’s decision boundary for the DoS class.

\subsubsection{SHAP results for LSTM}\label{ssubsection_LSTMSHAPperformance}
To evaluate the prediction behavior of the long short-term memory (LSTM) model, SHAP was again employed, utilizing the KernelExplainer for compatibility with the model’s sequential data structure. Since LSTM models require three-dimensional input, the test and background samples were first flattened into two-dimensional arrays to enable compatibility with SHAP's input expectations. A total of 100 samples were used for both the background dataset and the evaluation set to ensure computational tractability.

To enhance the interpretability of the LSTM model used for multiclass intrusion detection, SHAP (figure - \ref{fig:lstmshap} ) values were computed with respect to class 0, which represents Denial of Service (DoS) attacks. The SHAP summary plot illustrates the contribution of each input feature to the model’s decision to classify an instance as a DoS attack. Features are ranked along the Y-axis according to their average impact on model output, while the X-axis shows the distribution of SHAP values, indicating how strongly each feature pushes the prediction towards or away from the DoS class. The color gradient represents the original feature values, ranging from low (blue) to high (red).
The most influential feature identified was logged-in, where low values (blue) had strong positive SHAP values, suggesting that unauthenticated sessions significantly increased the likelihood of the model predicting a DoS attack. This aligns with known attack patterns, where DoS attempts typically do not involve legitimate logins. Conversely, features such as srv\_serror\_rate, dst\_host\_srv\_serror\_rate, and serror\_rate exhibited high SHAP values when their feature values were high (red), meaning that elevated server error rates are strong indicators of DoS activity. These findings reinforce the role of connection-based anomaly features in distinguishing malicious network behavior, particularly the high frequency of failed or rejected service requests that are symptomatic of DoS attacks.
Additional features such as protocol-type, count, and dst\_host\_serror\_rate also demonstrated moderate influence. For example, certain protocol types and connection counts affected the model’s confidence in detecting DoS attacks, reflecting the LSTM’s temporal sensitivity to sequence-level patterns in traffic. Meanwhile, lower-ranked features like wrong-fragment, rerror-rate, and hot had limited impact on the output, suggesting that these features carried less discriminative power for DoS detection in the LSTM model.

\subsection{User Experience Analysis}\label{subsection_UIperformance}

To comprehensively evaluate user perceptions of the proposed system, the User Interface (UI) component was designed to facilitate structured interaction and feedback collection across key psychosocial and experiential dimensions. Participants (15, 75\% male and 25\% female, Bangladeshi, specialist in the field of cybersecurity)  first provided basic demographic information, including age, gender, education level, and prior experience with intelligent systems. To complement this, a brief assessment of personality traits was conducted using the Mini-IPIP6~\cite{sibley2011mini} This allowed for an enriched understanding of how individual differences—across dimensions such as Honesty-Humility, Emotionality, Extraversion, Agreeableness, Conscientiousness, and Openness to Experience—may influence system perception, usability evaluation, and trust calibration~\cite{zhang2020effect}. Following this, participants engaged with a domain-relevant case study scenario to ground their experience in a realistic decision-making context. They then completed a structured post-interaction survey comprising three validated constructs: Trust, Reliability, and Usability. Trust was measured using items adapted from the short trust in automation scale (STIAS)~\cite{mcgrath2025monitoring}, reliability via the TiA framework~\cite{korber2018theoretical} and usability using items from the System Usability Scale (SUS)~\cite{brooke1996sus}. Each construct used a 5-point Likert scale, with scoring procedures adhering to standard guidelines.

\begin{table}[!h]
\centering
\caption{Consistency of the Validating Questions via Cronbach's Alpha Test}
\label{tab:UIperformance}
\begin{tabular}{|l|c|}
\hline
\textbf{User Experience Metrics} & \textbf{Cronbach's Alpha Score} \\
\hline
Trust                            & 0.90                            \\
\hline
Reliability                      & 0.90                            \\
\hline
System Usability                 & 0.60                            \\
\hline
\end{tabular}
\end{table}

\begin{figure}[htbp]
\centering
\includegraphics[scale=.25]{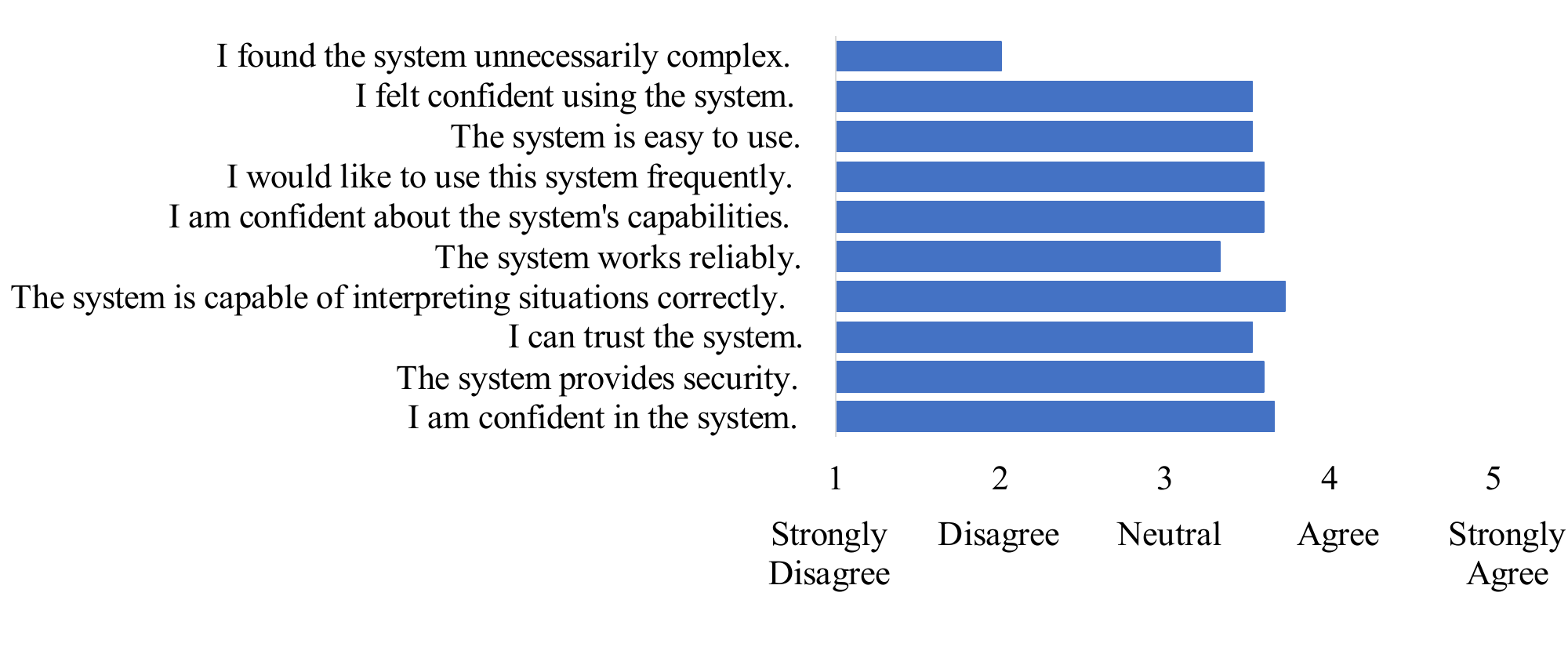}
\caption{Summary of user survey responses on a 5-point Likert scale}
\label{fig:likert}
\end{figure}

To evaluate internal consistency, Cronbach’s Alpha was calculated. Table \ref{tab:UIperformance} presents the overall consistency results of the Cronbach's Alpha test. The survey instruments demonstrated acceptable internal consistency (Cronbach’s, $\alpha$C = 0.70), with high reliability for trust ($\alpha$T  = 0.90) and reliability ($\alpha$R  = 0.90), and moderate consistency for usability ($\alpha$U  = 0.60), indicating an area for improvement. The results validate the use of adapted psychometric tools and underscore the role of psychosocial factors in shaping user experience. The use of a realistic case study scenario further enriched the feedback and contextual relevance. Figure \ref{fig:likert} displayed a summary of the UI-executed survey responses on a 5-point Likert scale.

\section{Conclusion \& Future Work}\label{section_conc_fut}
{
This study presents a network intrusion detection approach that enhances detection accuracy while maintaining interpretability through the use of SHAP. The results demonstrate that lightweight, interpretable models can be effective in low-resource environments, supporting practical deployment in real-world scenarios.
However, the trade-off between performance and interpretability remains a fundamental challenge. SHAP offers useful insights but may not fully capture deep model behavior. Future work will explore more advanced techniques, such as DeepLIFT~\cite{kalakoti2025evaluating} and Grad-CAM~\cite{sun2024novel}, to improve transparency. Additionally, due to hardware limitations, our experiments were constrained to simplified models and downsampled datasets. Expanding this work to larger datasets and more powerful computing environments will be key to validating its broader applicability. Also, future work will expand the sample size beyond the current 15 participants to enhance generalizability. We also aim to incorporate behavioral metrics to assess user reliance, complementing self-reported data. Usability and reliability will be improved through iterative, user-centered design, fostering greater trust and smoother human-system interaction.

\bibliographystyle{ieeetr}
\bibliography{template}

\end{document}